\documentclass{article}

\usepackage[preprint,nonatbib]{neurips_2024}

\usepackage[utf8]{inputenc} % allow utf-8 input
\usepackage[T1]{fontenc}    % use 8-bit T1 fonts
\usepackage{hyperref}       % hyperlinks
\usepackage{url}            % simple URL typesetting
\usepackage{booktabs}       % professional-quality tables
\usepackage{amsfonts}       % blackboard math symbols
\usepackage{nicefrac}       % compact symbols for 1/2, etc.
\usepackage{microtype}      % microtypography
\usepackage{xcolor}         % colors
\usepackage{algorithm}
\usepackage{algorithmic}
\usepackage{amsmath}

\usepackage{graphicx}
\usepackage{float}
\usepackage[numbers,sort&compress]{natbib}

\title{Scaling LLM Pre-training with \\ Vocabulary Curriculum}
\author{%
  Fangyuan Yu \\
  Temus\\
}

\begin{document}

\maketitle

\begin{abstract}

Modern language models rely on static vocabularies, fixed before pretraining, in contrast to the adaptive vocabulary acquisition observed in human language learning. To bridge this gap, we introduce vocabulary curriculum learning, an approach that improves pretraining efficiency with log-linear scaling gains relative to vocabulary size. Our method alternates between entropy-guided vocabulary expansion and model optimization, enabling models to learn transferable representations across diverse tokenization granularities. This approach naturally gives rise to an optimal computation allocation pattern: longer tokens capture predictable content, while shorter tokens focus on more complex, harder-to-predict contexts. Experiments on small-scale GPT models demonstrate improved scaling efficiency, reinforcing the effectiveness of dynamic tokenization. We release our code to support further research and plan to extend our experiments to larger models and diverse domains.

\end{abstract}

\section{Introduction}

\begin{figure}[H]
    \centering
    \includegraphics[width=0.9\textwidth]{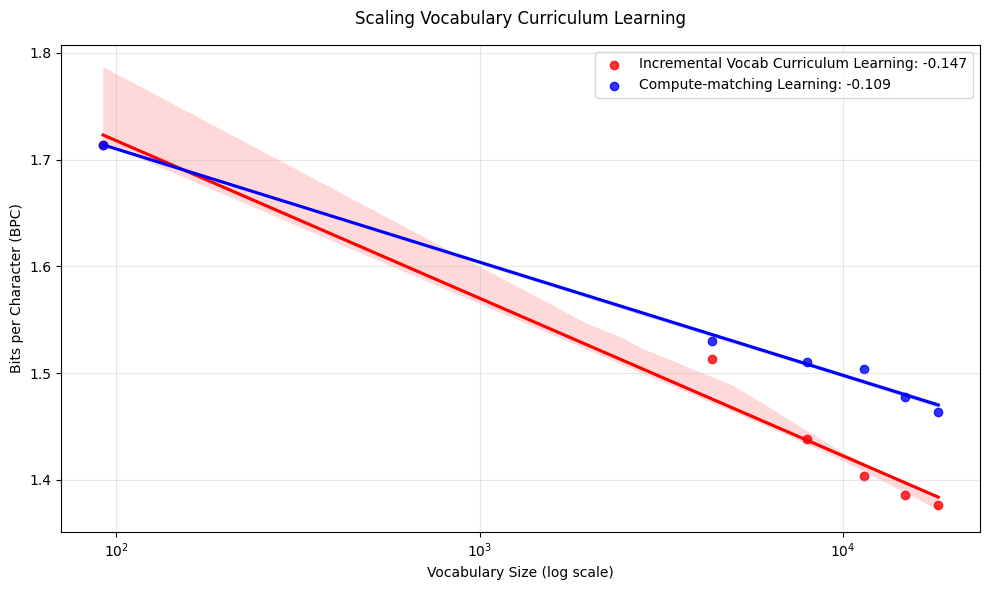}
    \caption{Scaling better with vocabulary curriculum}
    \label{fig:scaling_vocab}
\end{figure}
Modern language model pre-training relies on static vocabularies, fixed before training and detached from the model's learning dynamics—unlike human language acquisition. This fixed approach limits models' ability to adapt to different levels of linguistic granularity, potentially hindering efficiency and performance. While humans acquire language hierarchically, starting with basic units before building more complex representations, language models typically operate with predetermined tokenization schemes.

\begin{figure}[t]
    \centering
    \includegraphics[width=1.0\textwidth]{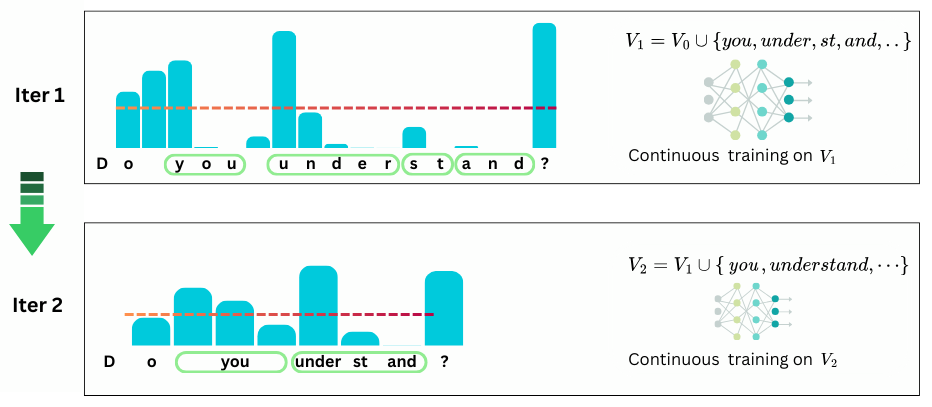}
    \caption{Scaling better with vocabulary curriculum}
    \label{fig:vocab_curriculum_main}
\end{figure}

Our approach dynamically merges predictable tokens, enabling the model to allocate computational resources more efficiently and shift focus toward harder-to-predict patterns. This results in an adaptive curriculum that evolves alongside the model's capabilities. The vocabulary curriculum learning strategy begins with basic units (characters) and progressively expands to more complex representations, allocating more capacity to regions of high modeling entropy and refining the model's understanding of difficult linguistic structures. A digram of our approach is provided in \ref{fig:vocab_curriculum_main}

Empirical results from pre-training GPT models \cite{radford2019gpt2} on the enwiki8 dataset \cite{hutter2006enwiki} highlight two key advantages of our vocabulary curriculum learning approach:

\begin{enumerate}
    \item It improves model performance across various vocabulary sizes, consistently achieving lower bits-per-character (BPC) compared to traditional fixed-vocabulary training.
    \item It enhances scaling efficiency—models trained with a vocabulary curriculum exhibit a shallower slope (0.109 vs. 0.147) in log-scale vocabulary size vs. BPC plots, indicating more effective utilization of larger vocabularies.
\end{enumerate}

As shown in Figure \ref{fig:scaling_vocab}, models trained with incremental vocabulary curriculum learning (red) exhibit a steeper improvement curve compared to compute-matching baselines (blue). The log-scale vocabulary size vs. bits-per-character (BPC) plot reveals that vocabulary curriculum learning achieves a slope of -0.147, meaning it leverages larger vocabularies more effectively than compute-matching learning, which only reaches -0.109.

Additionally, we observe that the curated vocabulary naturally forms a hierarchical structure, where longer tokens become increasingly predictable (lower BPC), while shorter tokens remain harder to predict (higher BPC). This structural organization emerges organically from our training process, reinforcing the effectiveness of our dynamic tokenization strategy.

Our key contributions are:

\begin{itemize}
    \item A dynamic vocabulary creation system that adapts based on model entropy
    \item A curriculum learning approach for tokenization that improves scaling efficiency
    \item Evidence that hierarchical token organization emerges naturally from our approach
\end{itemize}
While our focus is on language modeling, we believe this scaling effect can generalize to other modalities and domains, as byte sequences serve as the fundamental building blocks of digital data.

\section{Relevant Work}

\subsection{Tokenization Methods and Limitations}

Standard tokenization approaches like Byte Pair Encoding (BPE) \cite{gage1994bpe}, \cite{sennrich2016bpe} rely on static co-occurrence statistics detached from model learning. This creates representational limitations, particularly evident in early language models' struggles with mathematical operations \cite{radford2019gpt2}. Naive BPE tokenization produces inconsistent representations of numbers—for example, "711" might be encoded as a single token while "703" requires multiple tokens. This inconsistency makes it harder for models to learn arithmetic operations compared to specialized approaches that assign unique tokens to all 1-3 digit integers \cite{singh2024tokenizationcountsimpacttokenization}.

Even with a fixed vocabulary, different encoding strategies can produce varying segmentations of the same text. BPE-dropout \cite{provilkov2020bpedropoutsimpleeffectivesubword} leverages this property by introducing stochasticity during training, showing improvements in neural machine translation. More recent work exploits segmentation equivariance during inference to enhance reasoning through self-consistency \cite{sathe2024improvingselfconsistencyllms}. Additionally, research \cite{tao2024scalinglawsvocabularylarger} has established the existence of optimal vocabulary sizes for BPE-style tokenization, which correlate with model size, a log-linear relationship is observed between perplexity and vocabulary size. 

\subsection{Curriculum Learning}

Curriculum learning \cite{Bengio2009CurriculumL} progressively increases task difficulty during training to improve model performance. While successful in LLM post-training \cite{yu2024iterativegraphalignment, lee2024llm2llmboostingllmsnovel}, effective curriculum strategies for pre-training remain challenging \cite{campos2021curriculumlearninglanguagemodeling}. Previous attempts at vocabulary-based curricula for decoder-only models found no improvements \cite{feng2024childdirectedspeecheffectivetraining}, highlighting the difficulty of designing effective curricula for language model pre-training. Our work addresses these limitations with a novel adaptive approach to vocabulary curriculum.

\subsection{Entropy Aware Tokenization}

Recent work has begun exploring entropy-aware tokenization. The Byte Latent Transformer \cite{pagnoni2024bytelatenttransformerpatches} builds tokenization vocabularies using separately trained small language models. However, this approach creates vocabularies that are detached from the actual model's entropy patterns and cannot be dynamically updated during training.

Our work differs by integrating vocabulary building directly into the training process, allowing the tokenization scheme to evolve with the model's developing understanding of the text. This creates a true curriculum that adapts to the specific learning trajectory of each model, rather than imposing a static or pre-computed vocabulary structure.

\section{Approach}

Given a text corpus $\mathcal{D}$ consisting of numerous character sequences, where each sequence $x_{1:m} \in \mathcal{D}$ consists of characters $x_i$ (or bytes). A vocabulary $\mathcal{V}$ and an encoding function $e(x_{1:m}|\mathcal{V})$ together define a tokenization scheme that converts character sequences to token sequences $s_{1:n}$. Language modeling then focuses on predicting the next token: $p(s_n|s_{1:n-1})$ through minimizing entropy $H(s_{t}|s_{1:t-1})$.

We propose a dynamic tokenization framework that jointly learns the vocabulary and encoding strategy alongside the language model. Our approach consists of two key components: (1) entropy-guided vocabulary update and (2) vocabulary curriculum learning.

\subsection{Entropy-Guided Vocabulary Update}

Given a trained language model $f$, we identify mergeable token sequences based on their predictability. For a sequence $(s_1, s_2, \dots, s_n)$, we compute the entropy $H(s_t | s_{1:t-1})$ for each token. A sequence is considered mergeable if all tokens after the first position exhibit monotonically decreasing entropy below threshold $\epsilon$:

$$\text{mergeable}(s_{1:n}) \iff \forall t > 1: H(s_t | s_{1:t-1}) < H(s_{t-1} | s_{1:t-2}) \land H(s_t | s_{1:t-1}) < \epsilon$$

The vocabulary update process can either increase or decrease the vocabulary size:
$$\mathcal{V}_{k+1} = \begin{cases} 
\text{add}(\mathcal{V}_k, f, \mathcal{D}) & \text{for vocabulary expansion} \\
\text{reduce}(\mathcal{V}_k, n_{target}) & \text{for vocabulary reduction}
\end{cases}$$

where $n_{target}$ is the target vocabulary size.

For each new token added to $\mathcal{V}_{k+1}$, we expand the model's embedding layer $W_E \in \mathbb{R}^{|\mathcal{V}| \times d}$ and language modeling head $W_L \in \mathbb{R}^{|\mathcal{V}| \times d}$:
\begin{equation}
    W_E[v_{new}] = h_t^{(L)}, \quad W_L[v_{new}] = W_L[v_t]
\end{equation}
where $h_t^{(L)}$ represents the final hidden state for the merged sequence.

\begin{figure}[H]
    \centering
    \includegraphics[width=0.8\textwidth]{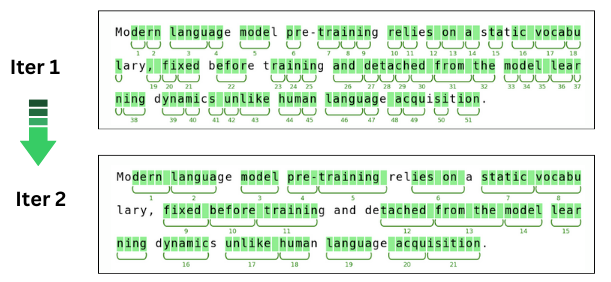}
    \caption{Token grouping process based on entropy patterns from a trained character-level language model}
    \label{fig:group_token}
\end{figure}

Unlike BPE which prohibits merges across space characters, our encoding function $e(x_{1:m}|\mathcal{V})$ allows unrestricted merging. The encoding process identifies longest valid token sequences using a sliding window approach, optimized through a trie structure for efficient prefix matching. To speed up encoding speed for long sequences, we employ batch encoding with additional tokenization at batch boundaries, considering connection sequences of length up to $\max_{v \in \mathcal{V}} |v|$.

The vocabulary management preserves several invariants: (1) non-leaf tokens are preserved during removal to maintain dependencies, (2) token indices reflect merge dependencies where tokens with smaller indices cannot be merges of tokens with larger indices, and (3) token indices align with rows in $W_E$ and $W_L$, enabling vocabulary reduction through prefix slicing.

\subsection{Vocabulary Curriculum Learning}

The curriculum learning process starts with the base vocabulary $\mathcal{V}_0 = \mathcal{A}$ (the character alphabet) and alternates between model optimization and vocabulary updates. At each stage $k$:

1. \textbf{Model Training}: Train the language model $f$ with current vocabulary $\mathcal{V}_k$ using cross-entropy loss:
   $$\mathcal{L}_k = -\sum_{x_{1:m} \in \mathcal{D}} \sum_{t} \log p(s_t | s_{1:t-1}; \mathcal{V}_k)$$
   where $s_{1:n} = e(x_{1:m}|\mathcal{V}_k)$ is the encoded sequence.

2. \textbf{Vocabulary Update}: Based on the trained model's entropy patterns, either expand the vocabulary through entropy-guided merging or reduce it through prefix slicing as defined in the previous section:
   $$\mathcal{V}_{k+1} = \begin{cases} 
   \text{add}(\mathcal{V}_k, f, \mathcal{D}) & \text{for expansion phase} \\
   \text{reduce}(\mathcal{V}_k, n_{target}) & \text{for reduction phase}
   \end{cases}$$

This iterative process continues until reaching the desired model performance or vocabulary size constraints.

\section{Experiments}

We investigate two key questions: (1) Is learning transferable across different vocabularies? and (2) Does vocabulary curriculum improve model performance? 

\subsection{Experimental Setup}

We evaluate our approach on a cleaned version of enwiki8 dataset using a small GPT architecture (context length 512, 6 layers, 6 attention heads, embedding dimension 384, ~10M parameters). The initial vocabulary $\mathcal{V}_0$ consists of 92 characters, and the model is trained with dropout 0.2 without bias terms. For vocabulary updates, we set the entropy threshold $\epsilon=0.3$ and limit per-iteration vocabulary growth to 3K tokens. 

\subsection{Incremental Vocabulary Curriculum}

Our primary experiment consists of 5 iterations of vocabulary expansion, starting from a base model with minimal vocabulary (92) and progressively training models with larger vocabularies (4359, 7941, 11382, 14819, 18276). Each iteration uses the previous model's checkpoint for vocabulary addition. We cap the vocabulary at 18K based on compute-matching experiments showing performance deterioration beyond this size, aligning with observations in \cite{tao2024scalinglawsvocabularylarger} that optimal vocabulary size correlates with model size.

\begin{figure}[H]
    \centering
    \includegraphics[width=1.0\textwidth]{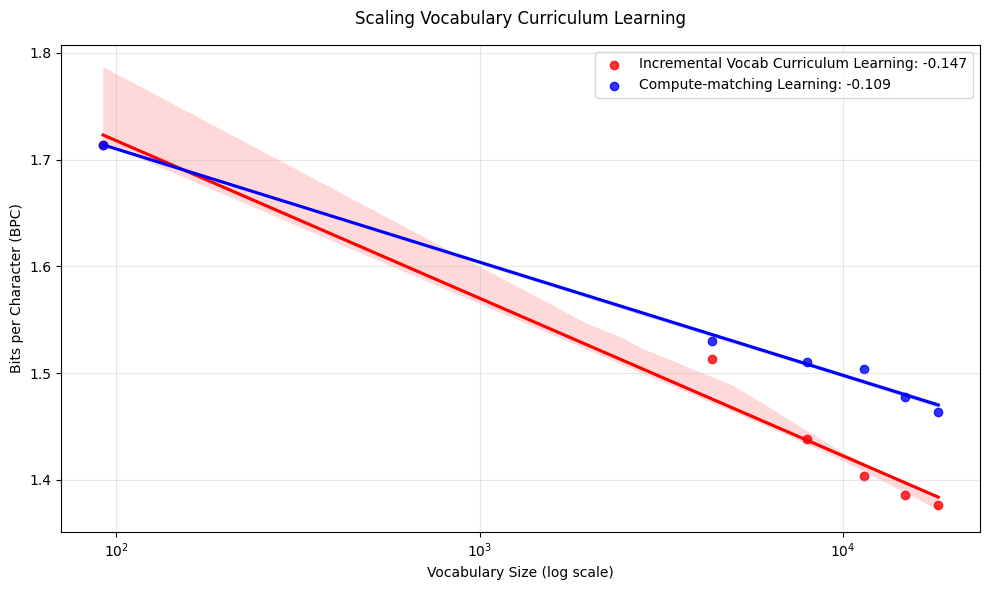}
    \caption{Incremental vocabulary learning shows noticeable improvement which scales with vocabulary size in log-linear fashion}
    \label{fig:scaling_vocab1}
\end{figure}

To isolate curriculum effects from training duration, we compare against compute-matching baselines where models are trained from scratch with equivalent total iterations. As shown in Figure \ref{fig:scaling_vocab1}, training with progressively increasing vocabulary reveals a log-linear relationship between vocabulary size and Bits Per Character (BPC), with curriculum learning demonstrating a steeper improvement curve compared to baseline training, detailed BPC at each iteration is documented in \ref{tab:bpc_comparison}

\begin{table}[htbp]
  \centering
  \caption{Comparison of BPC Values Across Different Vocabulary Sizes}
  \label{tab:bpc_comparison}
  \begin{tabular}{lrrrrrr}
    \toprule
    \textbf{Method} & \multicolumn{6}{c}{\textbf{Vocabulary Size}} \\
    \cmidrule(lr){2-7}
    & \textbf{92} & \textbf{4,359} & \textbf{7,941} & \textbf{11,382} & \textbf{14,819} & \textbf{18,276} \\
    \midrule
    incre\_vocab\_curriculum & 1.7141 & 1.5131 & 1.4385 & 1.4032 & 1.3853 & 1.3764 \\
    compute\_matching & 1.7141 & 1.5303 & 1.5103 & 1.5035 & 1.4780 & 1.4637 \\
    \midrule
    \% Improvement & 0.00\% & 1.12\% & 4.75\% & 6.67\% & 6.27\% & 5.96\% \\
    \bottomrule
  \end{tabular}
\end{table}

\subsection{Analysis of Improvement Mechanisms}

\begin{figure}
    \centering
    \includegraphics[width=1.0\textwidth]{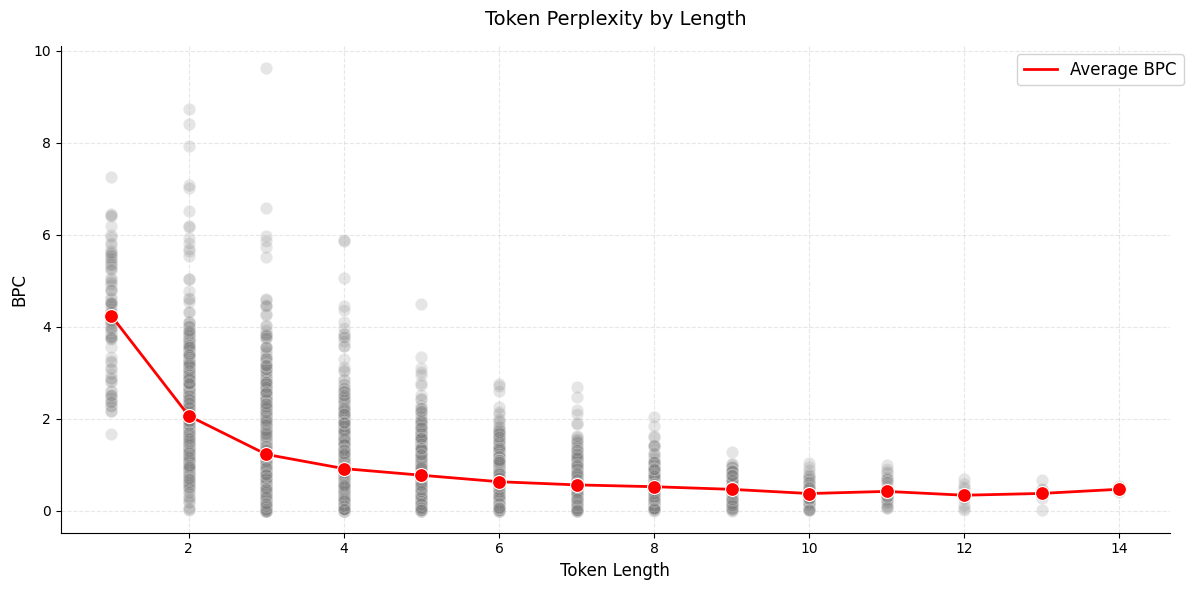}
    \caption{Longer tokens has smaller BPC, contributing to smaller global BPC}
    \label{fig:iter1_bpc_token_len}
\end{figure}

To understand the source of these improvements, we analyze per-token BPC distributions across different checkpoints. Figure \ref{fig:iter1_bpc_token_len} shows that at vocabulary size 4359, longer tokens consistently achieve better compression rates, validating our entropy-aware token addition approach.

\begin{figure}
    \centering
    \includegraphics[width=1.0\textwidth]{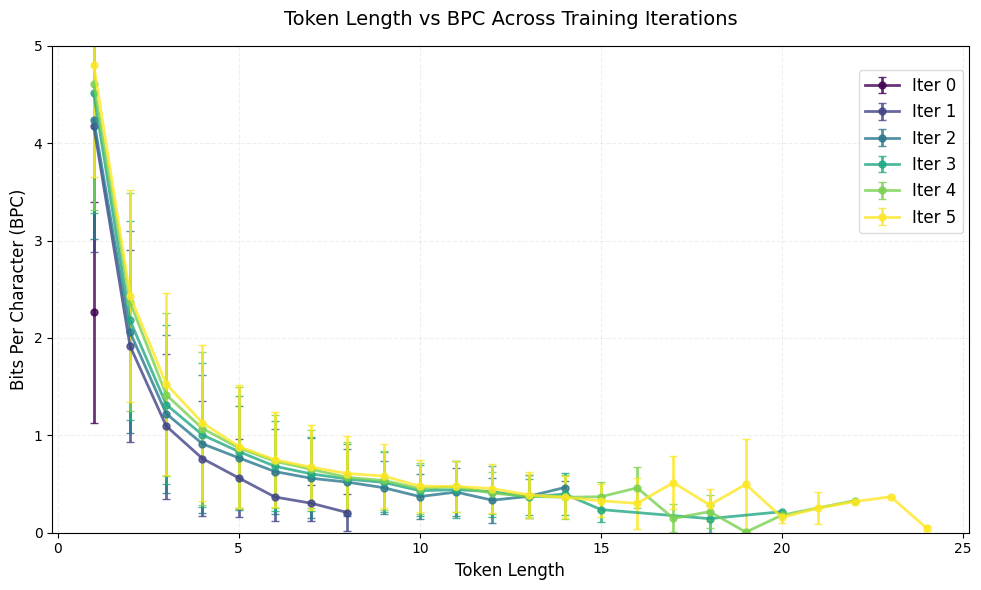}
    \caption{Longer tokens has smaller BPC, contributing to smaller global BPC}
    \label{fig:bpc-per-len-across-iter}
\end{figure}

Further analysis across iterations (Figure \ref{fig:bpc-per-len-across-iter} and Table \ref{tab:bpc_per_iter}) reveals two key patterns:
1. Newly created tokens are progressively longer and achieve lower BPC
2. Original shorter tokens become more challenging to model, showing slight BPC increases

\begin{table}[h]
\centering
\begin{tabular}{cccccccc}
\hline
\textbf{Token Group} & \textbf{Iter 0} & \textbf{Iter 1} & \textbf{Iter 2} & \textbf{Iter 3} & \textbf{Iter 4} & \textbf{Iter 5} \\
\hline
Iter 0 tokens & 2.26 & 4.20 & 4.24 & 4.47 & 4.60 & 4.82 \\
Iter 1 tokens & - & 1.11 & 1.22 & 1.30 & 1.41 & 1.47 \\
Iter 2 tokens & - & - & 0.78 & 0.83 & 0.88 & 0.91 \\
Iter 3 tokens & - & - & - & 0.77 & 0.80 & 0.84 \\
Iter 4 tokens & - & - & - & - & 0.76 & 0.80 \\
Iter 5 tokens & - & - & - & - & - & 0.76 \\
\hline
\end{tabular}
\caption{BPC values across training iterations. Each row represents tokens introduced at a specific iteration, while columns show how their BPC values change in subsequent iterations. Note the increasing BPC trend for early tokens (top rows) and lower initial BPC for tokens introduced later (bottom rows).}
\label{tab:bpc_per_iter}
\end{table}

This suggests that our curriculum enables the model to effectively learn hierarchical patterns, with longer tokens capturing predictable sequences while shorter tokens specialize in harder-to-predict contexts.

Notably, when testing a decremental vocabulary curriculum (reducing vocabulary size over time), we observe performance comparable to direct training but without additional improvements, suggesting that the benefits of curriculum learning are specifically tied to the incremental approach. 

\subsection{Implication and Future work}

Optimal vocabulary size is correlated with model size \cite{tao2024scalinglawsvocabularylarger}, following this insight, we suspect the scaling improvement might be better for bigger model size. We'll work on extending our experiments therein.  The effectiveness of incremental vocabulary learning suggests its potential application in other modality than text, for instance, in bGPT \cite{wu2024languagemodelsbytemodels}  all digital files can be converted into byte sequences, where the scaling power of vocabulary curriculum could be leveraged to compress the context, as well as improve modeling accuracy.


\begin{thebibliography}{99}

\bibitem{martinez2023climbcurriculumlearninginfantinspired}
R.~D. Martinez, Z.~Goriely, H.~McGovern, C.~Davis, A.~Caines, P.~Buttery, and L.~Beinborn, ``CLIMB: Curriculum Learning for Infant-inspired Model Building,'' \textit{arXiv preprint arXiv:2311.08886}, 2023.

\bibitem{provilkov2020bpedropoutsimpleeffectivesubword}
I.~Provilkov, D.~Emelianenko, and E.~Voita, ``BPE-Dropout: Simple and Effective Subword Regularization,'' \textit{Association for Computational Linguistics}, pp. 1882--1892, 2020.

\bibitem{sathe2024improvingselfconsistencyllms}
A.~Sathe, D.~Aggarwal, and S.~Sitaram, ``Improving Self Consistency in LLMs through Probabilistic Tokenization,'' \textit{ICML2024}, 2024.

\bibitem{pagnoni2024bytelatenttransformerpatches}
A.~Pagnoni, R.~Pasunuru, P.~Rodriguez, J.~Nguyen, B.~Muller, M.~Li, C.~Zhou, L.~Yu, J.~Weston, L.~Zettlemoyer, G.~Ghosh, M.~Lewis, A.~Holtzman, and S.~Iyer, ``Byte Latent Transformer: Patches Scale Better Than Tokens,'' \textit{arXiv: 2412.09871}, 2024.

\bibitem{tao2024scalinglawsvocabularylarger}
C.~Tao, Q.~Liu, L.~Dou, N.~Muennighoff, Z.~Wan, P.~Luo, M.~Lin, and N.~Wong, ``Scaling Laws with Vocabulary: Larger Models Deserve Larger Vocabularies,'' \textit{NeurIPS 2024}, 2024.

\bibitem{tian2024tokenizeworldobjectlevelknowledge}
R.~Tian, B.~Li, X.~Weng, Y.~Chen, E.~Schmerling, Y.~Wang, B.~Ivanovic, and M.~Pavone, ``Tokenize the World into Object-level Knowledge to Address Long-tail Events in Autonomous Driving,'' \textit{arXiv preprint arXiv:2407.00959}, 2024.

\bibitem{singh2024tokenizationcountsimpacttokenization}
A.~K. Singh and D.~J. Strouse, ``Tokenization counts: the impact of tokenization on arithmetic in frontier LLMs,'' \textit{arXiv: 2402.14903}, 2024.

\bibitem{gage1994bpe}
P.~Gage, ``A New Algorithm for Data Compression,'' \textit{The C Users Journal}, 1994.

\bibitem{sennrich2016bpe}
R.~Sennrich, B.~Haddow, and A.~Birch, ``Neural Machine Translation of Rare Words with Subword Units,'' in \textit{Proceedings of the 54th Annual Meeting of the Association for Computational Linguistics (ACL)}, pp. 1715--1725, 2016.

\bibitem{radford2019gpt2}
A.~Radford, J.~Wu, R.~Child, D.~Luan, D.~Amodei, and I.~Sutskever, ``Language Models are Unsupervised Multitask Learners,'' \textit{OpenAI Blog}, 2019.

\bibitem{hutter2006enwiki}
M.~Hutter, ``The Human Knowledge Compression Prize,'' 2006.

\bibitem{wu2024languagemodelsbytemodels}
S.~Wu, X.~Tan, Z.~Wang, R.~Wang, X.~Li, and M.~Sun, ``Beyond Language Models: Byte Models are Digital World Simulators,'' \textit{arXiv: 2402.19155}, 2024.

\bibitem{Bengio2009CurriculumL}
Y.~Bengio, J.~Louradour, R.~Collobert, and J.~Weston, ``Curriculum learning,'' in \textit{International Conference on Machine Learning}, 2009.

\bibitem{lin2025rho1tokensneed}
Z.~Lin, Z.~Gou, Y.~Gong, X.~Liu, Y.~Shen, R.~Xu, C.~Lin, Y.~Yang, J.~Jiao, N.~Duan, and W.~Chen, ``Rho-1: Not All Tokens Are What You Need,'' \textit{arXiv:2404.07965}, 2025.

\bibitem{yu2024iterativegraphalignment}
F.~Yu, H.~S. Arora, and M.~Johnson, ``Iterative Graph Alignment,'' \textit{arXiv:2408.16667}, 2024.

\bibitem{campos2021curriculumlearninglanguagemodeling}
D.~Campos, ``Curriculum learning for language modeling,'' \textit{arXiv:2108.02170}, 2021.

\bibitem{warstadt2023papersbabylmchallenge}
A.~Warstadt, L.~Choshen, A.~Mueller, A.~Williams, E.~Wilcox, and C.~Zhuang, ``Call for Papers -- The BabyLM Challenge: Sample-efficient pretraining on a developmentally plausible corpus,'' \textit{arXiv: 2301.11796}, 2023.

\bibitem{chollet2019measureintelligence}
F.~Chollet, ``On the Measure of Intelligence,'' \textit{arXiv preprint arXiv:1911.01547}, 2019.

\bibitem{feng2024childdirectedspeecheffectivetraining}
S.~Y. Feng, N.~D. Goodman, and M.~C. Frank, ``Is Child-Directed Speech Effective Training Data for Language Models?,'' \textit{arXiv preprint arXiv:2408.03617}, 2024.

\bibitem{chen2024squidlongcontextnew}
W.~Chen, Z.~Li, S.~Xin, and Y.~Wang, ``Squid: Long Context as a New Modality for Energy-Efficient On-Device Language Models,'' \textit{arXiv preprint arXiv:2408.15518}, 2024.

\bibitem{lee2024llm2llmboostingllmsnovel}
N.~Lee, T.~Wattanawong, S.~Kim, K.~Mangalam, S.~Shen, G.~Anumanchipalli, M.~W. Mahoney, K.~Keutzer, and A.~Gholami, ``LLM2LLM: Boosting LLMs with Novel Iterative Data Enhancement,'' \textit{arXiv preprint arXiv:2403.15042}, 2024.

\end{thebibliography}
\end{document}